\documentclass{article}

\usepackage{amsmath,amsfonts,bm}

\def\eqref#1{equation~\ref{#1}}
\def\1{\bm{1}}

\DeclareMathAlphabet{\mathsfit}{\encodingdefault}{\sfdefault}{m}{sl}
\SetMathAlphabet{\mathsfit}{bold}{\encodingdefault}{\sfdefault}{bx}{n}

\newcommand{\R}{\mathbb{R}}

\usepackage{PRIMEarxiv}

\usepackage[utf8]{inputenc} % allow utf-8 input
\usepackage[T1]{fontenc}    % use 8-bit T1 fonts
\usepackage{hyperref}       % hyperlinks
\usepackage{url}            % simple URL typesetting
\usepackage{booktabs}       % professional-quality tables
\usepackage{amsfonts}       % blackboard math symbols
\usepackage{nicefrac}       % compact symbols for 1/2, etc.
\usepackage{microtype}      % microtypography
\usepackage{lipsum}
\usepackage{fancyhdr}       % header
\usepackage{graphicx}       % graphics
\graphicspath{{media/}}     % organize your images and other figures under media/ folder
\usepackage{amssymb}
\usepackage{latexsym}
\usepackage{wrapfig}

\usepackage{graphicx}
\usepackage{mathtools}
\usepackage{amsfonts}
\usepackage{pifont}
\usepackage{booktabs}
\usepackage{todonotes}
\usepackage{siunitx}
\usepackage{microtype}
\usepackage{dirtytalk}
\usepackage{amsthm}
\usepackage{makecell} % for more than one line in a tabular
\usepackage{xcolor}
\usepackage[ruled,linesnumbered]{algorithm2e}
\usepackage{subcaption}
\usepackage{multirow}
\usepackage{placeins}
\usepackage{threeparttable}

\SetKwInput{KwInput}{Input}
\SetKwInput{KwOutput}{Output}
\DeclarePairedDelimiterX{\inner}[2]{\langle}{\rangle}{#1, #2}

\newcounter{ArasCounter}

\newcounter{GittaCounter}

\newcounter{HolgerCounter}

\newcounter{PhilippCounter}

\DeclareMathOperator{\sgn}{sgn}

\newcommand*\widebar[1]{\@ifnextchar^{{\wide@bar{#1}{0}}}{\wide@bar{#1}{1}}}

\title{Interpretable Robotic Friction Learning via Symbolic Regression
}

\author{
  Philipp Scholl \\
  Ludwig-Maximilians-Universität München \\
  Munich \\
  Germany\\
  \texttt{scholl@math.lmu.de} \\
     \And
  Alexander Dietrich \\
  German Aerospace Center \\
  Weßling \\
  Germany\\
     \And
  Sebastian Wolf \\
  German Aerospace Center \\
  Weßling \\
  Germany\\
     \And
  Jinoh Lee \\
  German Aerospace Center \\
  Weßling \\
  Germany\\
     \And
  Alin Albu-Schäffer \\
  German Aerospace Center \\
  Weßling \\
  Germany\\
     \And
  Gitta Kutyniok \\
  Ludwig-Maximilians-Universität München \\
  German Aerospace Center \\
  University of Troms\o{} \\
  Munich Center for Machine Learning (MCML) \\
  Munich \\
  Germany\\
     \And
  Maged Iskandar \\
  German Aerospace Center \\
  Weßling \\
  Germany\\
}

\begin{document}
\maketitle

\begin{abstract}
Accurately modeling the friction torque in robotic joints has long been challenging due to the request for a robust mathematical description. Traditional model-based approaches are often labor-intensive, requiring extensive experiments and expert knowledge, and they are difficult to adapt to new scenarios and dependencies. On the other hand, data-driven methods based on neural networks are easier to implement but often lack robustness, interpretability, and trustworthiness\textemdash{}key considerations for robotic hardware and safety-critical applications such as human-robot interaction. To address the limitations of both approaches, we propose the use of symbolic regression (SR) to estimate the friction torque. SR generates interpretable symbolic formulas similar to those produced by model-based methods while being flexible to accommodate various dynamic effects and dependencies. In this work, we apply SR algorithms to approximate the friction torque using collected data from a KUKA LWR-IV+ robot. Our results show that SR not only yields formulas with comparable complexity to model-based approaches but also achieves higher accuracy. Moreover, SR-derived formulas can be seamlessly extended to include load dependencies and other dynamic factors.
\end{abstract}

% keywords can be removed

\section{Introduction}

The estimation of the joint friction torque in robots is crucial for various tasks, including force control \citep{lakatos2013modally,iskandar2023hybrid}, safe physical human-robot interaction \citep{de2008atlas,iskandar2024intrinsic,vogel2020ecosystem,iskandar2021}, and external torque estimation \citep{liu2021sensorless}. 
Modeling friction presents significant challenges due to its dependence on multiple factors. This complexity limits the practicality of model-based approaches, as they often require expert knowledge, considerable time, and varied methodologies to accommodate different dynamics. Additionally, incorporating new dependencies, such as load or temperature, further complicates the process \citep{Linderoth2013, bittencourt2012static, bittencourt2013modeling}. 

This has prompted many research efforts to turn to data-driven methods, such as neural networks \citep{Selmic2002, Ciliz2007, Huang2012, Guo2019, scholl2024friction}. In \citet{Selmic2002} a specialized neural network architecture for friction using discontinuous activations is introduced, which improves data fit while minimizing neurons, training time, and data requirements. 
Hybrid approaches that combine simple parametric models with complex neural networks have also gained traction.
The superiority of these models has been demonstrated in \citet{Ciliz2007}, leveraging the flexibility of neural networks and the inductive bias from known dynamics. Expanding on this, \citet{Huang2012} developed dual neural networks for friction, while \citet{Guo2019} combined individual networks for inertia, Coriolis, and gravitational torque into a unified model, also employing discontinuous activation functions. Furthermore, \citet{scholl2024friction} introduced a second network to adapt friction estimators to new dynamics efficiently.

While data-driven models such as neural networks can outperform model-based approaches and easily adapt to new scenarios, their black-box nature raises concerns about trust and robustness. 
Symbolic regression (SR) \citep{la2021contemporary} is a machine learning technique that derives mathematical expressions directly from data without relying on a predefined model structure. This makes SR particularly useful to describe physical phenomena that lack explicit mathematical formulations or are too complex to derive analytically.  Moreover, SR can uncover new, interpretable relationships hidden in the data, making it well-suited for approximating complex behaviors like friction torque, which is difficult to express precisely due to its dependence on multiple interacting factors.
However, existing SR applications to friction in robotic joints suffer from limitations. The only attempt, to the best of our knowledge is \citet{sandi2023determining}, which uses second-hand data extracted from a figure, introducing significant noise. Moreover, the resulting formula is complex, difficult to interpret, and it lacks the precision required for practical implementations, as acknowledged by the authors.

In this paper, we focus on developing interpretable and accurate friction models using SR, which also provides flexibility in incorporating additional factors.
To achieve this, we collect detailed friction data from a KUKA LWR-IV+ robot.  In contrast to \citet{sandi2023determining}, we demonstrate that SR algorithms can outperform conventional friction models. Furthermore, we showcase the ease of applying such an SR approach to new scenarios by incorporating load effects and variations.
Three different SR approaches are used and evaluated for various scenarios. The considered algorithms consistently produce interpretable formulas that significantly outperform model-based approaches and prove useful for estimating the external torque of the robot joints.

The rest of the article is organized as follows. Sec.~\ref{sec:robot-dynamics} introduces background on the robot dynamics and model-based friction estimation. In Sec.~\ref{sec:symbolic-regression} we give an introduction into SR and state-of-the-art algorithms from this field, which we will use in our experiments later. The datasets are described in Sec.~\ref{sec:data}. Sec.~\ref{sec:experimental-results} shows our experimental results. Finally, Sec.~\ref{sec:conclusion} concludes the paper.

\section{BACKGROUND} \label{sec:robot-dynamics}

This section outlines the robot dynamics and detail the assumptions made during training, which aim to isolate the friction effects. The dynamics of the robot can be described as
\begin{equation} 
\bm{M}(\bm{q}) \bm{\Ddot{q}} + \bm{C}(\bm{q},\bm{\Dot{q}}) \bm{\Dot{q}} + \bm{\tau}_g(\bm{q}) = \bm{\tau}_m + \bm{\tau}_f + \bm{\tau}_{ext}, 
\end{equation}
where $\bf q$, $\bf \Dot{q}$, and $\bf \Ddot{q}$ represent the joint position, velocity, and acceleration vectors, respectively, and $\bf M(q)$ is the positive definite inertia matrix. The term $\bf C(q,\Dot{q})$ corresponds to the Coriolis and centrifugal matrix, $\bm{\tau}_g(\bm{q})$ represents the gravitational torque, and $\bm{\tau}_f$ is the friction torque. Additionally, $\bm{\tau}_m$ and $\bm{\tau}_{ext}$ represent the motor joint torque and external joint torque, respectively.
Here, full knowledge of the dynamic components $(\bm{M}(\bm{q}), \bm{C}(\bm{q},\bm{\Dot{q}}), \bm{\tau}_g(\bm{q}))$ is assumed, while $\bm{\tau}_f$ is unknown. Furthermore, $\bm{\tau}_m$ is computed directly through the motor current.

In the case where no external force is applied $( \bm{\tau}_{ext} = \bm{0})$, the motor torque can be expressed as a function of the known robot dynamics and friction torque as follows:

\begin{equation} 
\bm{\tau}_m = \bm{M}(\bm{q}) \bm{\Ddot{q}} + \bm{C}(\bm{q},\bm{\Dot{q}}) \bm{\Dot{q}} + \bm{\tau}_g(\bm{q}) - \bm{\tau}_f. 
\end{equation}
In the specific scenario where constant, single-joint velocities or low, near-constant velocities are assumed ${(\bf \Ddot{q}\approx 0, C(q,\Dot{q}) \Dot{q}\approx0)}$, the quadratic terms of individual joints become negligible, and the coupling terms are considered zero. This simplifies the dynamics to

\begin{equation} \label{eq:friction-low-constant-velocity} 
\bm{\tau}_m = \bm{\tau}_g(\bm{q}) - \bm{\tau}_f. 
\end{equation}

These assumptions are applied during training, allowing us to relate the motor torque to the friction effects as described by (\ref{eq:friction-low-constant-velocity}).
The friction torque $\bm{\tau}_f$ can be modeled by incorporating several key characteristics that describe friction in the sliding regime, including static friction, Coulomb friction, viscous friction, and the Stribeck effect \citep{armstrong2012}. 

In the sliding regime, static friction $\tau_{\text{f,s}}$ can be expressed through various functional forms, with a common model (the LuGre Model \citep{Canudas1995}) \footnote{Note that (\ref{eq:s_friction_model}) is a scalar equation, since the friction is an individual joint-level effect.}  given as
\begin{equation}\begin{split}\tau_{\text{f,s}}(\dot{q})&=g(\dot{q})+s(\dot{q})\\ g(\dot{q})&=\sgn(\dot{q})\left(F_{\text{c}}+(F_{\text{s}}- F_{\text{c}})e^{-\vert \dot{q}/ v_{\text{s}}\vert ^{\delta_{\text{s}}}}\right),\end{split}
\label{eq:s_friction_model}
\end{equation}
where, $s(\dot{q})$ represents the velocity-strengthening term known as viscous friction. Typically, this is linearly proportional to the joint velocity $\dot{q}$, where $s(\dot{q}) = F_v \dot{q}$ and $F_v$ is a constant coefficient. In general, the velocity-strengthening term may also include nonlinear components, as discussed in \citet{iskandar2019}.
The function $g(\dot{q})$ captures the velocity-weakening behavior of static friction, also known as the Stribeck curve, as it models the Stribeck effect. In this case, $F_{\text{c}}$ represents Coulomb friction, $F_{\text{s}}$ denotes static (or stiction) friction, $v_{\text{s}}$ is the Stribeck velocity, and $\delta_{\text{s}}$ is the exponent governing the nonlinearity of the Stribeck effect.

%In the Gaussian parametrization as used in [19] and [20] the exponent parameter is δs = 2. Parameters of the total friction torque τr,s can be identified easily using the static map between the friction torque and the relative velocity.
The LuGre model has multiple possible extensions. As the friction phenomenon is, by nature, nonlinear and continuous at zero velocity crossing, it can be of a disadvantage to use the static friction model which is discontinuous at velocity reversal. Therefore, some works \citep{Canudas1995,johanastrom2008revisiting} extend the LuGre model to a continuous dynamical systems and incorporate load dependency. Furthermore, friction of a Harmonic-Drive (HD) gear-based robotic joint is known to be highly dependent on the temperature, which can be incorporated in the static and dynamic friction models \citep{iskandar2019, wolf2018}. 
While dynamic friction models show high accuracy in capturing the physical friction effects of the robotic joints, the difficulty of estimating and adapting their parameters remains. This limits the usage of such models in many scenarios and proves a strong motivation for data-driven approaches.

\section{SYMBOLIC REGRESSION} \label{sec:symbolic-regression}

Symbolic regression (SR) \citep{la2021contemporary} is a machine learning technique that seeks to find an accurate mathematical model for a dataset without relying on predefined assumptions about its structure. The goal is to identify an interpretable, symbolic function $f:\R^n \rightarrow \R$ that best fits the data points $(x_i, y_i)_{i=1,\dots,N} \subseteq \R^n \times \R$, such that $y_i \approx f(x_i)$ for all $i$. In general, the user provides a set of base functions\textemdash{}such as $\exp$, $\sin$, or $\log$\textemdash{}that the SR algorithm is permitted to use, thereby offering some control over the resulting formula. Traditionally, SR methods utilize genetic programming to perform a heuristic search over the space of possible functions \citep{augusto2000symbolic, cranmer2023interpretable}. More recently, reinforcement learning has been employed to find the simplest and most accurate formulas \citep{petersen2019deep} and has also been combined with genetic programming \citep{mundhenk2021symbolic}.
In another approach, \citet{udrescu2020tegmark} addresses the problem by recursively simplifying it using symmetries commonly found in physical laws. Other techniques translate the discrete optimization problem into a continuous one by representing the function space as a linear combination of basis functions \citep{brunton2016} or complex compositions of standard base functions \citep{martius2016extrapolation,scholl2023parfam}.

More recently, transformers have been applied to this problem, utilizing pre-training on synthetic data to predict symbolic formulas directly \citep{kamienny2022end} or combine it with reinforcement learning \citep{Landajuela2022usdr}.
In this work, we select the most competitive methods from each subfield \citep{cranmer2023interpretable, scholl2023parfam, Landajuela2022usdr}, as described in the following subsections.

\subsection{Genetic programming based symbolic regression} \label{sec:pysr}

\citet{cranmer2023interpretable} introduced PySR as a highly optimized library for SR using genetic programming. Genetic programming SR algorithms consider a function as a genome consisting of variables, constants, standard operations ($+, -,\cdot,/$), and a pre-defined set of base functions.A population of genomes is evolved through random mutations, crossovers, and survival of the fittest\textemdash{}eliminating genomes, i.e., functions, that poorly fit the data or are excessively complex.
PySR employs a multi-population evolutionary algorithm based on tournament selection \citep{brindle1980genetic} in the subpopulations. To avoid early convergence, age-regularized evolution \citep{real2019regularized} is incorporated, which works by replacing the oldest member of the subpopulations instead of the weakest member in the early phase of training. Furthermore, \citet{cranmer2023interpretable} introduces simulated annealing \citep{bertsimas1993simulated} to control the acceptance rate of tournament selection, therefore, controlling the diversity of the population. Simultaneously to the evolution, the equations are simplified using algebraic identities, and parameters are optimized using classical optimizers to identify equations with real constants. Additionally, PySR penalizes the complexity of functions on a population level, to encourage the algorithm to explore all complexity levels similarly. 

\subsection{Continuous optimization based symbolic regression} \label{sec:parfam}

ParFam \citep{scholl2023parfam} transforms the discrete optimization problem into a continuous one to accelerate the search using continuous optimization algorithms. This is achieved by spanning the space of "relevant" functions using a parametric family of functions that resemble a residual neural network \citep{he2016deep} with one hidden layer, with rational functions instead of linear functions as connections between the layers and physically meaningful activation functions, which are the base functions chosen by the user. This can be written compactly as 
\begin{equation} \label{eq:parfam}
    {f_\theta(x)\! = \! Q_{k+1}(x, g_1(Q_1(x)),g_2(Q_2(x)), \dots, g_k(Q_k(x))),}
\end{equation}

\noindent where $g_1,...,g_k$ are the base functions, $Q_1,...,Q_{k+1}$ are rational functions, $x$ is the input, and the learnable parameters $\theta$ are the coefficients of the polynomials of $Q_i$. To recover an interpretable formula from (\ref{eq:parfam}), the loss consists of the mean squared error and the 1-norm of $\theta$. It is optimized using the global search algorithm basin-hopping \citep{Wales1997GlobalOB}
%in combination with BFGS \citep{NW2006}
and a fine-tuning routine to increase the sparsity of $\theta$ and, thus, decrease the complexity of the learned function.

\subsection{Neural network based symbolic regression} \label{sec:udsr}

Unified deep symbolic regression (uDSR)~\citep{Landajuela2022usdr} is a combination of several SR algorithms: genetic programming, reinforcement learning, recursive simplification, pre-training, and linear models. Recursive problem simplification (AIFeynman) \citep{udrescu2020tegmark} is used to simplify the data. While \citet{udrescu2020tegmark} use brute-force or polynomial fitting to resolve the simplified problems, \citet{Landajuela2022usdr} apply their hybrid method. The hybrid method uses as a base deep symbolic regression (DSR)~\citep{petersen2019deep}, which trains a recurrent neural network %\citep{rumelhart1986learning} 
with reinforcement learning on one problem set to predict the correct equations. Samples from DSR serve then as the initial population for a standard genetic programming algorithm, similar to \citep{mundhenk2021symbolic}. This process is refined by including a \textit{linear} token, which represents expressions that are linear with respect to some chosen feature space, often the set of monomials with bounded degrees. The advantage of the token is its ability to quickly learn complex expressions that are typically challenging for methods like genetic programming. The last adaption is to pre-train the neural network employed by DSR on generic equations, instead of initializing the weights randomly.

\begin{figure}[t]
\centering
\begin{minipage}[c][0.24\textheight][c]{0.65\columnwidth}
    \centering
    \includegraphics[width=\linewidth, height=0.22\textheight, keepaspectratio]{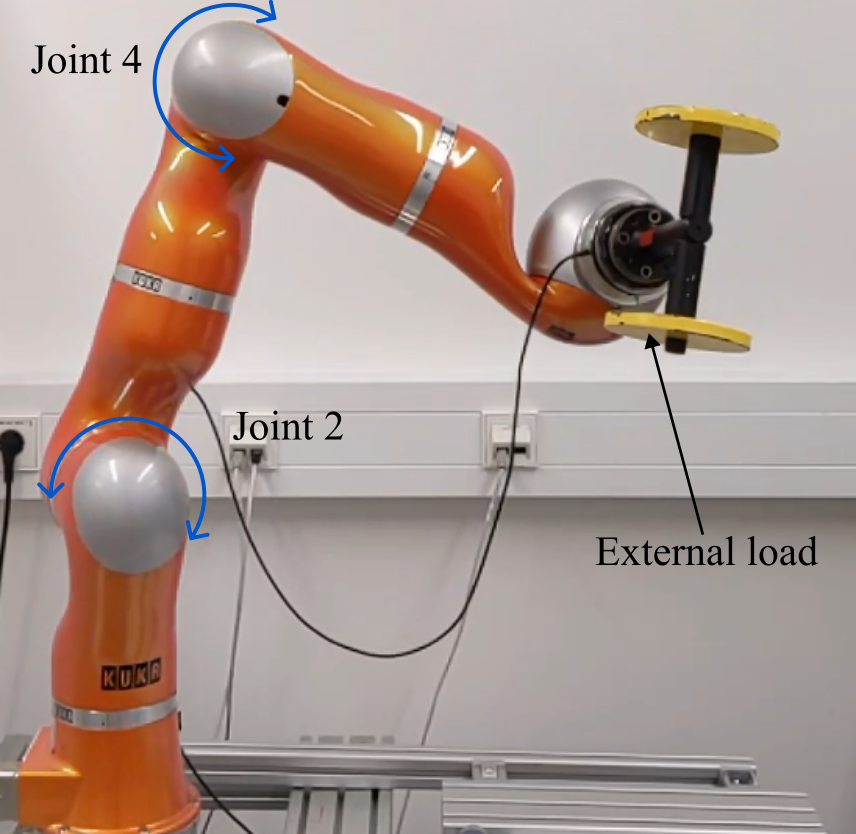}
\end{minipage}
\caption{The experimental setup: the robot is equipped with link-side torque sensors for the reference signals, while the measured motor current is used in the proposed method.}
\label{fig:robot}
\end{figure}

\begin{figure}[b]
    \centering
% \begin{minipage}[c][0.24\textheight][c]{0.95\columnwidth}
    \centering
    \includegraphics[width=0.95\columnwidth, height=0.22\textheight, keepaspectratio]{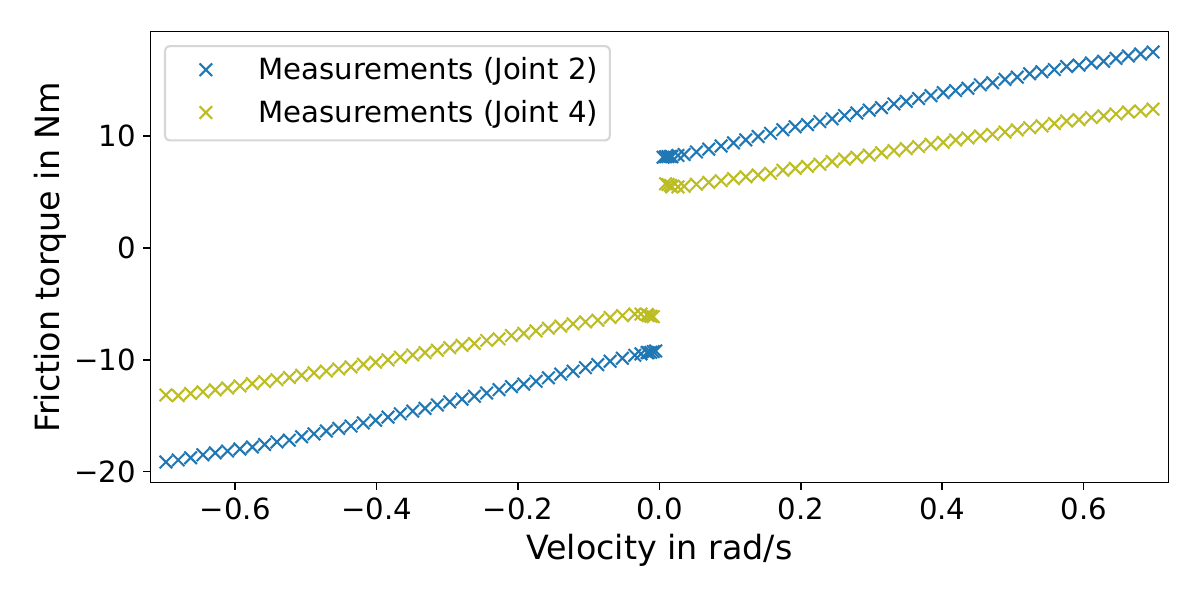}
% \end{minipage}
    \caption{The measured friction torque-velocity behavior for Joint 2 and Joint 4 of the KUKA LWR-IV+ robot for the base data set.}
    \label{fig:measurements-velocity-dependence}
\end{figure}

\section{Data Collection} \label{sec:data}

The data are collected to capture the physical behavior of the friction torque as accurately as possible. The experimental data are recorded from each robot joint individually at various constant velocities. For this research, we utilize the torque-controlled KUKA LWR-IV+ robot as the reference platform, focusing on the second and fourth joints (pitch axes), as these joints exhibit greater variability in friction effects due to the pronounced influence of the gravitational torque due to the change of joint configuration. The complete experimental setup is depicted in Fig.~\ref{fig:robot}.

The primary objective of this work is to model friction effects primarily arising from the HD gear. Although the robot is equipped with link-side torque sensors, these sensors are used exclusively for validation purposes. The data is divided into two main subsets, which are designed to best capture the friction behavior while minimizing the impact of unmodeled dynamics.

\subsection{Dataset A} \label{sec:static-data-set}

The first dataset acquisition is designed to capture the static friction behavior. To achieve this, the robot joints were subjected to single-joint constant velocities, ensuring that (\ref{eq:friction-low-constant-velocity}) holds, which allows for the direct computation of friction torque. Extensive experimental measurements were performed to cover the full operational velocity and position ranges for each joint. This approach enables a detailed characterization of static friction. By averaging over the friction for a constant velocity, the relationship between static friction and velocity can be isolated from other effects at a high resolution, as illustrated in Fig.~\ref{fig:measurements-velocity-dependence}.

\subsection{Dataset B} \label{sec:dynamic-data-set}

\begin{figure}[t]
    \centering
    \includegraphics[width=0.48\linewidth, keepaspectratio]{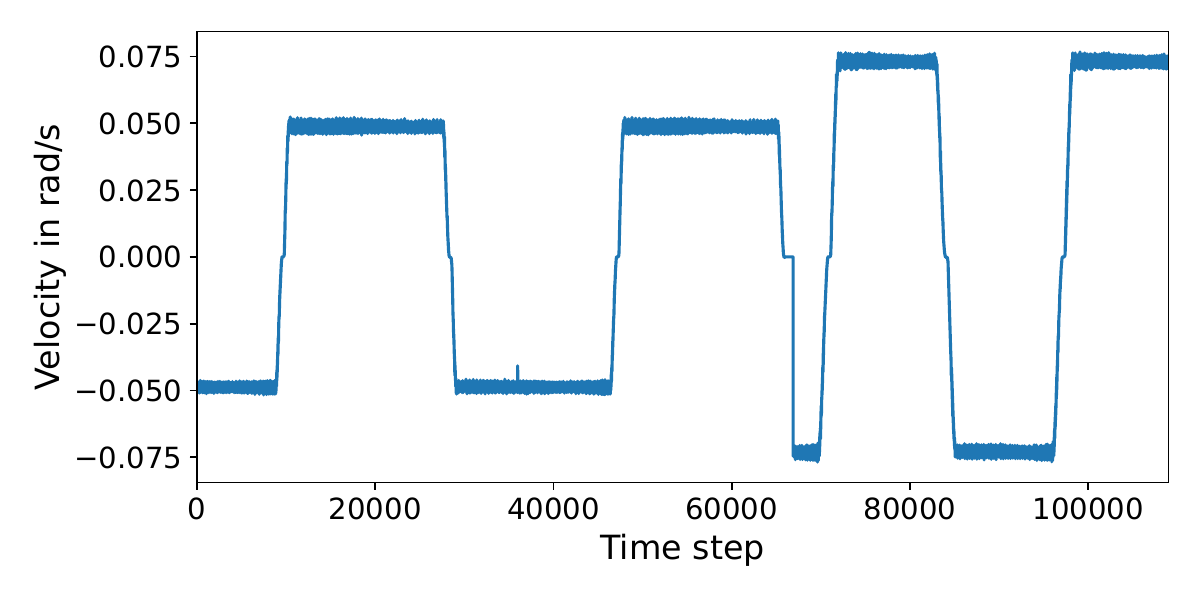}
    \includegraphics[width=0.48\linewidth, keepaspectratio]{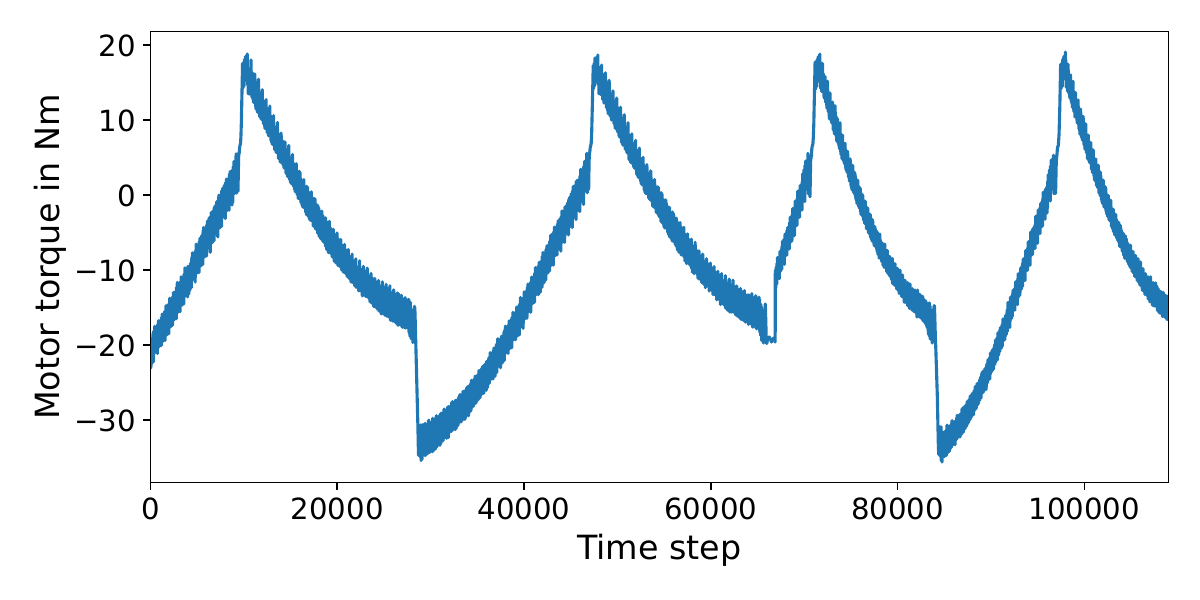}
     \caption{Plots of velocity and motor torque of Dataset B without external load for Joint 2, measured with a sampling rate of 1kHz.}
     \label{fig:small-dataset-plots}
\end{figure}

Dataset B is collected dynamically by applying different velocities sequentially while different known loads are attached to the robot end-effector and the joints are moved simultaneously. The velocity and motor torque for this dataset without external load/torque are displayed in Fig.~\ref{fig:small-dataset-plots}. By keeping the velocity low and constant we ensure that (\ref{eq:friction-low-constant-velocity}) can be used to compute the friction torque. Due to the dependence of the friction torque on the sign of the velocity and the load, this data set has different friction dynamics than Dataset A.%, see also \citep{scholl2024friction}.

\section{EXPERIMENTAL RESULTS} \label{sec:experimental-results}

In this section, we apply the SR algorithms PySR~\citep{cranmer2023interpretable}, ParFam \citep{scholl2023parfam}, and uDSR~\citep{Landajuela2022usdr} to the datasets introduced in Sec.~\ref{sec:data}. Throughout the experiments, we choose the exponential function ($\exp$) and the root function ($\sqrt{}$) as the base functions. As binary operations, we only allow addition, subtraction, and multiplication, since including the division operator for PySR resulted in formulas that were accurate on the given data but had mathematical singularities in between the measurements. The fact that this problem could easily be found without extensive numerical tests underlines the advantage of producing interpretable formulas. The computation time was below two minutes for each problem for each algorithm on a 11th Gen Intel® Core™ i7-1165G7 @ 2.80GHz × 8.

The static-friction model shown in (\ref{eq:s_friction_model}) is used as a base-line. Since the SR algorithms are not constrained to symmetric models, we also adopt an asymmetric model-based approach which consists of (\ref{eq:s_friction_model}) for two different sets of parameters $F_v$, $F_c$, $F_s$, $v_s$, and $\delta_s$ depending on the sign of $\dot{q}$. The parameters are computed using the SciPy \citep{2020SciPy-NMeth} implementation of the global optimizer basin-hopping \citep{Wales1997GlobalOB}. 

\subsection{Static friction without load dependency} \label{sec:exp-static-dataset}

As a first experiment, we use Dataset A and isolate the effect of the load on the friction torque and velocity by averaging over different gravitational torques, as shown in Fig.~\ref{fig:measurements-velocity-dependence}. Since we know that the sign of the velocity is an important component in this relation, we allow the SR algorithms to take both $\dot{q}$ and $\sgn(\dot{q})$ as inputs.% ParFam runs 16s and returns the following formula:

% \begin{equation} \label{eq:dataset-a-wo-load-dependency-parfam}
%     \begin{split}
%         \hat{\tau}_f = & \left(3.66 \dot{q} + 8.52 \sgn(\dot{q}) - 0.44\right) \cdot \\ 
%         & \exp{(- 0.78 \dot{q}^{2} + 1.26 |\dot{q}| + 0.02 \dot{q} - 0.02 \sgn(\dot{q}))}
%     \end{split}
% \end{equation}

% \philipp{get rid of the computation times (just mention all are below a few minutes)}

\begin{figure}[t]
    \centering
    % \begin{minipage}[c][0.64\textheight][c]{0.95\columnwidth}
        \centering
        \subfloat[\label{fig:dataset-a-joint-2} Joint 2]{%
            \includegraphics[width=\linewidth, height=0.3\textheight, keepaspectratio]{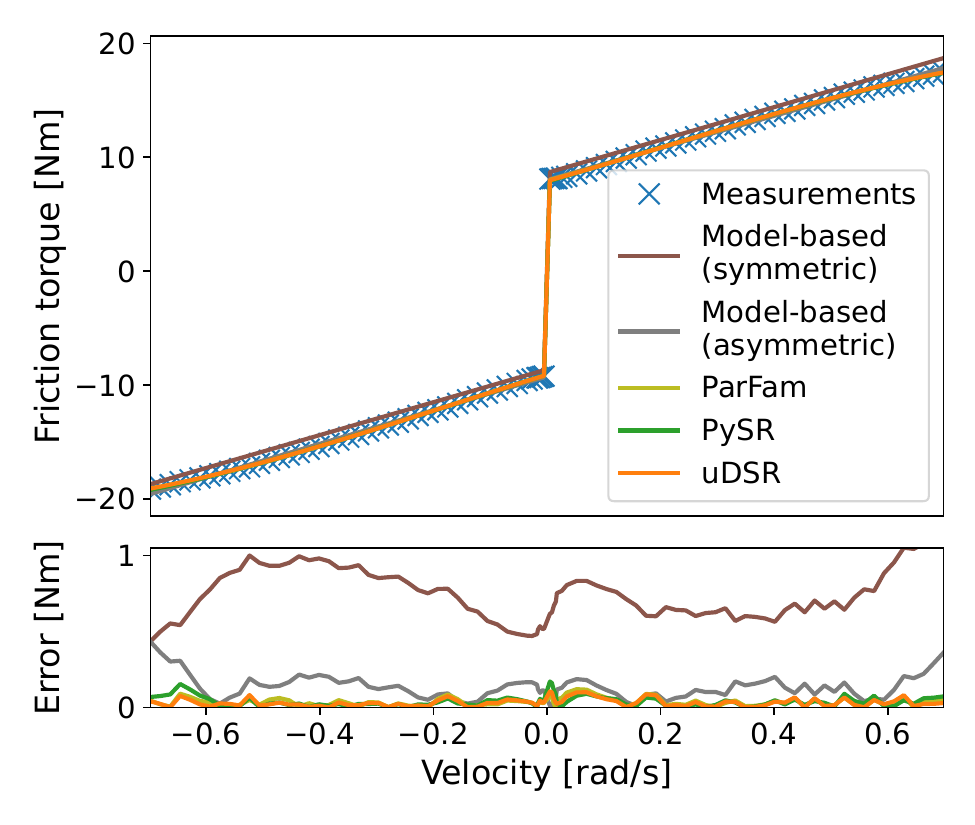}} % \\[0.5ex]
        \subfloat[\label{fig:dataset-a-joint-4} Joint 4]{%
            \includegraphics[width=\linewidth, height=0.3\textheight, keepaspectratio]{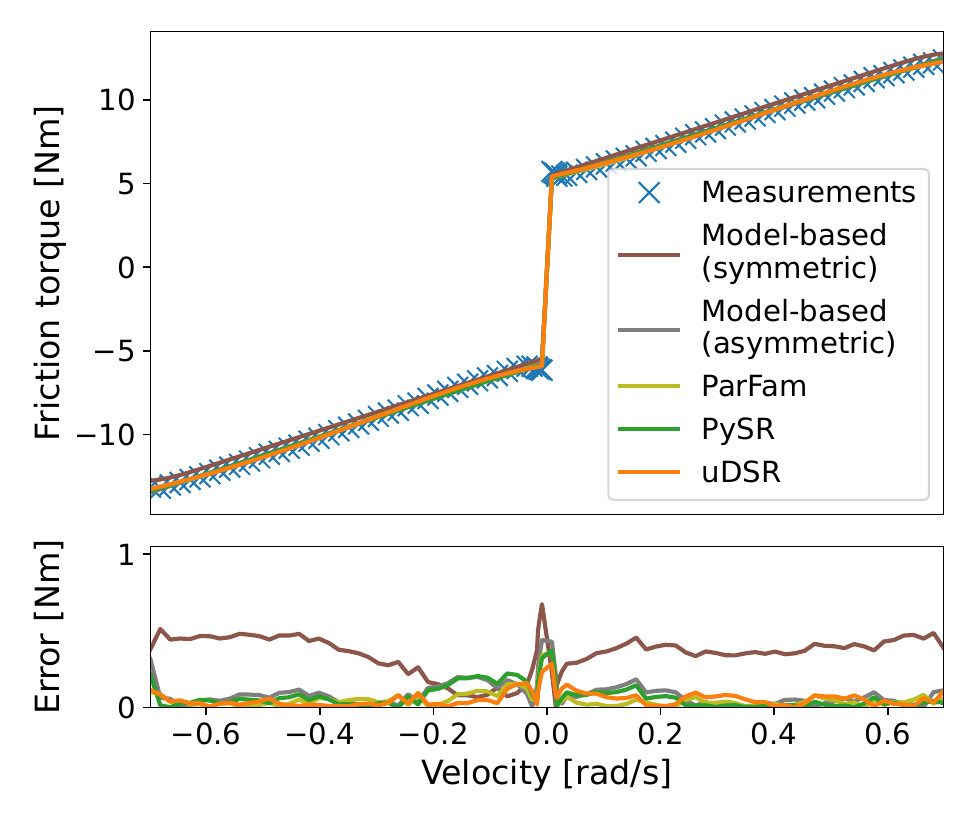}}
    % \end{minipage}
    \caption{Friction estimation of (a) Joint 2 and (b) Joint 4 of the KUKA robot using Dataset A for the friction-velocity relationship, where the top diagram shows the friction torque vs. joint velocity while the bottom diagram depicts the error w.r.t.~the considered range of velocity.}
    \label{fig:dataset-a-joint-2-4}
    \vspace{-0.3cm}
\end{figure}

% \begin{figure}[t]
%     \centering
%     \subfloat[\label{fig:dataset-a-joint-2} Joint 2]{\includegraphics[width=0.95\columnwidth]{Images/dataset_a_no_load_joint_2.pdf}}\\
%     \subfloat[\label{fig:dataset-a-joint-4} Joint 4]{\includegraphics[width=0.95\linewidth]{Images/dataset_a_no_load_joint_4.pdf}}
%     \caption{Friction estimation of (a) Joint 2 and (b) Joint 4 of the KUKA robot using Dataset A for the friction-velocity relationship, where the top diagram shows the friction torque vs. joint velocity while the bottom diagram depicts the error w.r.t.~the considered range of velocity.}
%     \label{fig:dataset-a-joint-2-4}
% \end{figure}

\begin{table*}
    % \begin{minipage}[c][0.29\textheight][c]{2\columnwidth}
    \scriptsize
    \centering
    \caption{Results on Dataset A without load dependency: friction estimation, complexity, and the computed formular $\hat{\tau}_f$ }
\label{tab:dataset-a-wo-load-dependency}
    \begin{threeparttable}
    \begin{tabular}{|c|c|c|c|c|c|}
    \hline
        \multirow{2}{*}{} & \multicolumn{2}{c|}{Estimation error [Nm]} & \multicolumn{2}{c|}{Complexity [-]} & $\hat{\tau}_f$ \\ %\hline
        & Joint 2 & Joint 4 & Joint 2 & Joint 4 & Joint 2\\ \hline          
        \makecell{Model-based \\ (symmetric)} & 0.737 & 0.350 & 20 & 20 & \makecell[l]{
        $\begin{aligned}
            \hat{\tau}_f= & \sgn(\dot{q})\left(1193+(8.629 - 1193)e^{-\vert \dot{q}/ 47.65\vert ^{8.827}}\right)+ 14.44 \dot{q}
        \end{aligned}$} \\ \hline
        \makecell{Model-based \\ (asymmetric)}& 0.134 & 0.093 & 40 & 40 & \makecell[l]{
        $\hat{\tau}_f=\begin{cases}
            \sgn(\dot{q})\left(264.3+(8.002- 264.3)e^{-\vert \dot{q}/ 95.33\vert ^{92.98}}\right)+14.10\dot{q}, & \dot{q}>0 \\
            \sgn(\dot{q})\left(773.6+(8.629- 773.6)e^{-\vert \dot{q}/ 69.23\vert ^{52.20}}\right)+14.44\dot{q}, & \dot{q}\leq0
        \end{cases}$
        }\\ \hline
        ParFam & 0.037 & 0.058 & 22 & 23 &         \makecell[l]{
            $\begin{aligned}
                \hat{\tau}_f = & \left(3.66 \dot{q} + 8.52 \sgn(\dot{q}) - 0.44\right) \cdot \\
                &\exp{(- 0.78 \dot{q}^{2}+ 1.26 |\dot{q}| + 0.02 \dot{q} - 0.02 \sgn(\dot{q}))}
            \end{aligned}$
        }
        \\ \hline
        PySR & 0.041 & 0.074 & 15 & 11 & \makecell[l]{$\begin{aligned}
            \hat{\tau}_f = & - 4.283 \dot{q}^{4} \sgn(\dot{q})\\ 
            &  + \left(8.452 - 0.215 \sgn(\dot{q})\right) \left(1.842 \dot{q} + \sgn(\dot{q}) - 0.049\right) 
            \end{aligned}$} 
        \\ \hline
        uDSR & 0.033 & 0.059 & 36 & 43 & \makecell[l]{$\begin{aligned}
        \hat{\tau}_f = & - 8.083 \dot{q}^{3} + 4.643 \dot{q}^{2} \sgn(\dot{q}) - 0.068 \dot{q}^{2} - 0.639 |\dot{q}| + \\ & 13.609 \dot{q} + 0.827 \sgn(\dot{q}) + 0.827 \sgn(\dot{q}) + e^{\dot{q}} + e^{e^{\sgn(\dot{q})}} - 9.899
        \end{aligned}$}
        \\ 
        \hline
    \end{tabular}
    
        \begin{tablenotes}
        \footnotesize
        \item[*] The estimation error denotes the mean absolute error of the friction estimation in Nm.
        \item[*] For brevity, we present the complexity of the computed formulas for Joints~2 and 4, and the computed formula for $\hat{\tau}_f$ for Joint~2.
        \end{tablenotes}
    \end{threeparttable}
    % \end{minipage}
    % \vspace{-0.4cm}
\end{table*}

\begin{figure}
    % \vspace{-1cm}
    \centering
    \subfloat[][Velocity input]{\includegraphics[width=0.5\columnwidth]{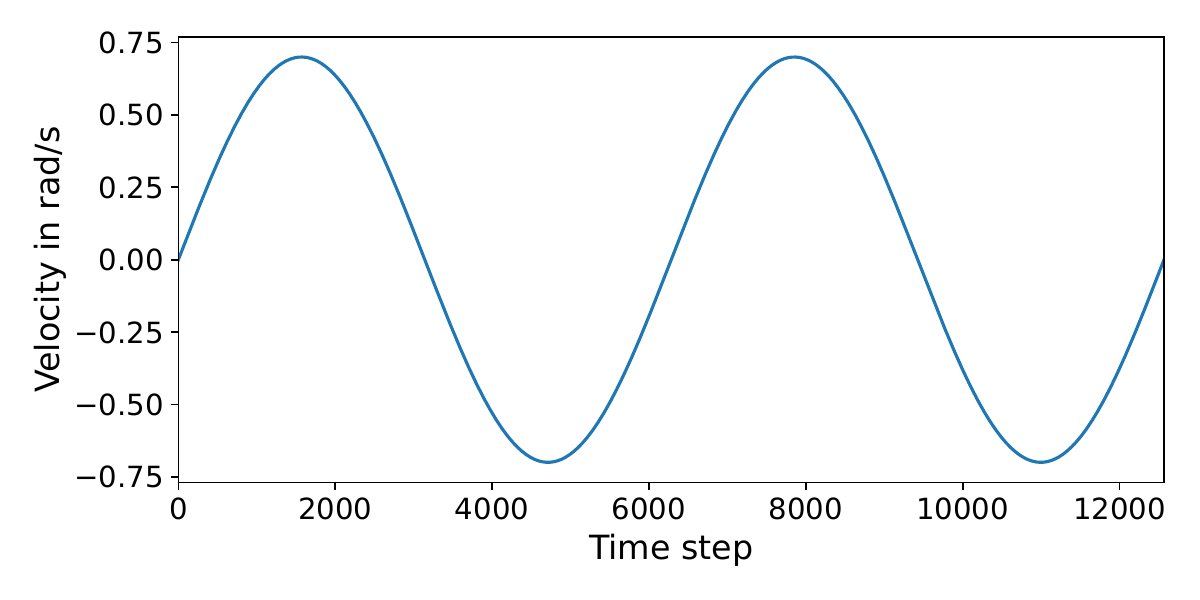}}\\
    % \caption{Asymmetric load} \label{fig:velocity-sinusoidal}
    \subfloat[Predicted friction torque at Joint 2]{\includegraphics[width=0.5\columnwidth]{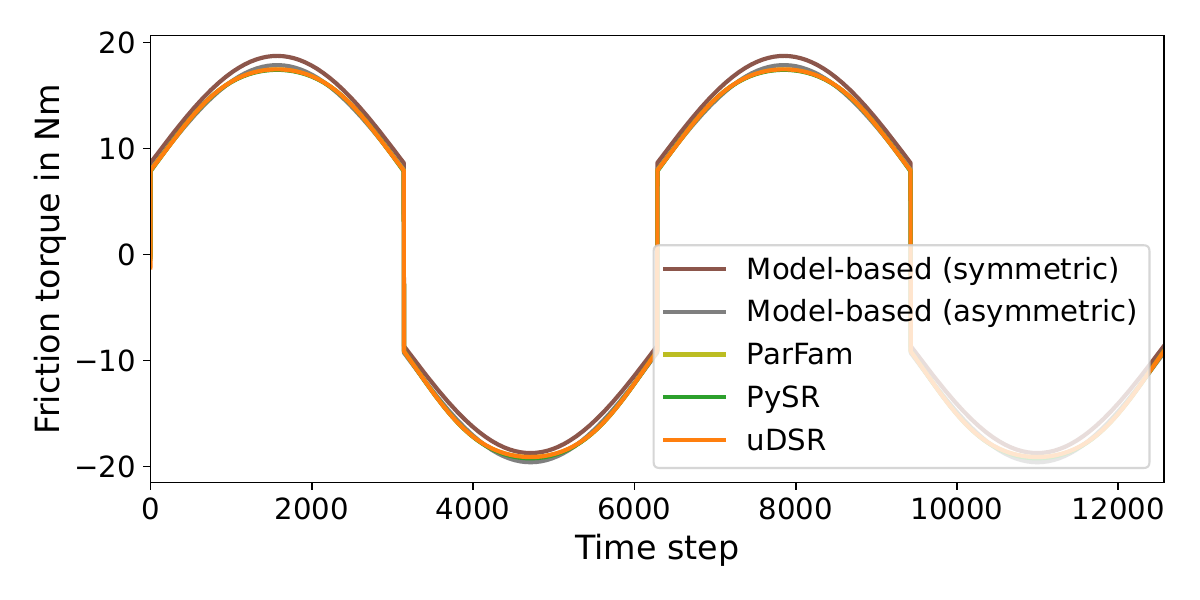}}
    % \caption{Symmetric load} \label{fig:friction-sinusoidal}
    \subfloat[Predicted friction torque at Joint 4]{\includegraphics[width=0.5\columnwidth]{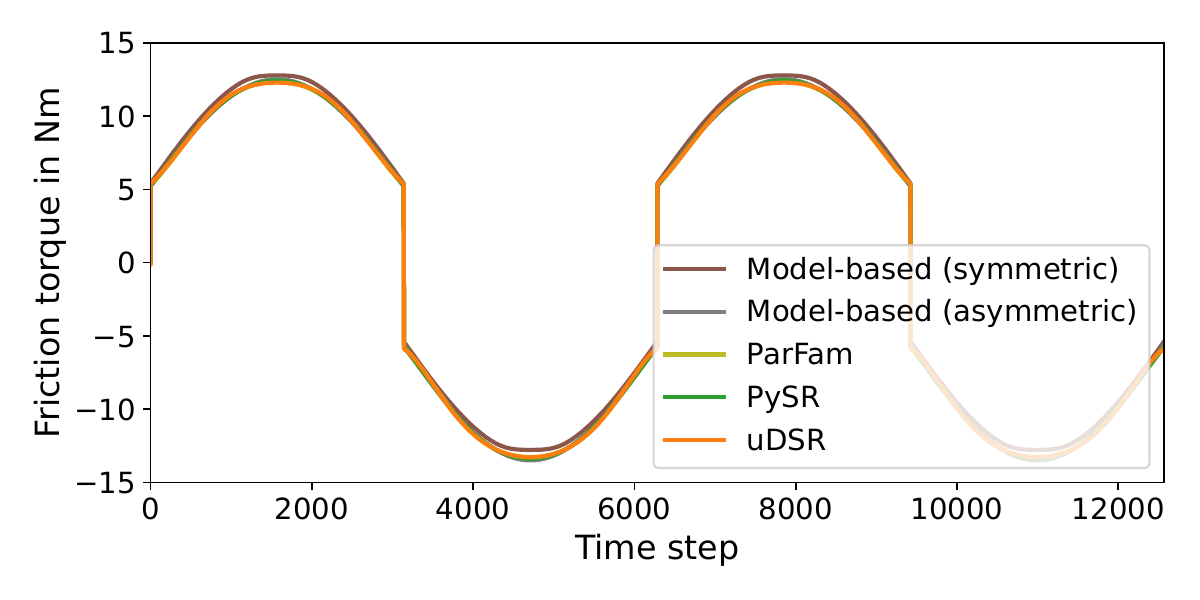}}
    % \caption{Symmetric load} \label{fig:friction-sinusoidal}
    \caption{Friction torque prediction for (a) sinusoidal velocity input is shown for (b) Joint 2 and (c) Joint 4, respectively.}
    \label{fig:dataset-a-sinusoidal}
\end{figure}

Fig.~\ref{fig:dataset-a-joint-2-4} and Table~\ref{tab:dataset-a-wo-load-dependency} show that the SR algorithms capture the friction torque better than the symmetric and even the asymmetric model. Table~\ref{tab:dataset-a-wo-load-dependency} also includes the learned formulas and their complexity, which is a standard metric used in SR \citep{la2021contemporary} that counts the number of operations ($+$, $-$, $\cdot$ , $/$), functions ($\sqrt{}$, $\exp$, potences), and variable accesses. Interestingly, all SR algorithms outperform both model-based approaches even though the found formulas of ParFam and PySR are arguably less complicated than the asymmetric form of (\ref{eq:s_friction_model}). To get a better understanding of the predictions, we plot the predictions of the five estimators for a pre-defined sinusoidal velocity profile for both joints in Fig.~\ref{fig:dataset-a-sinusoidal}. % Interestingly, this shows that PySR learned a singularity around zero velocity as can also be seen from its formula in Table~\ref{tab:dataset-a-wo-load-dependency}, an insight that would not have been possible with neural networks.
The lowest overall mean absolute error is achieved by uDSR with 0.033Nm and 0.059Nm for Joint 2 and 4, respectively. However, its formula is slightly more complicated than the ones found by ParFam and PySR. For the results of the other algorithms and of the exact formulas found, see Table~\ref{tab:dataset-a-wo-load-dependency}.

\subsection{Static friction with load dependency} \label{sec:exp-static-dataset-load-dependency}

While incorporating load dependency into model-based friction estimators is not straightforward, we demonstrate in the next step that this can be easily achieved using SR. To this end, we utilize the raw Dataset A, where the effect of the load on the friction torque is not isolated. This effect can be seen by the non-constant motor torque during a constant velocity motion as illustrated in Fig.~\ref{fig:small-dataset-plots}.  

\begin{figure}[t]
    \centering
    % \begin{minipage}[c][0.62\textheight][c]{0.95\columnwidth}
    \subfloat[Joint 2]{\includegraphics[width=0.5\linewidth]{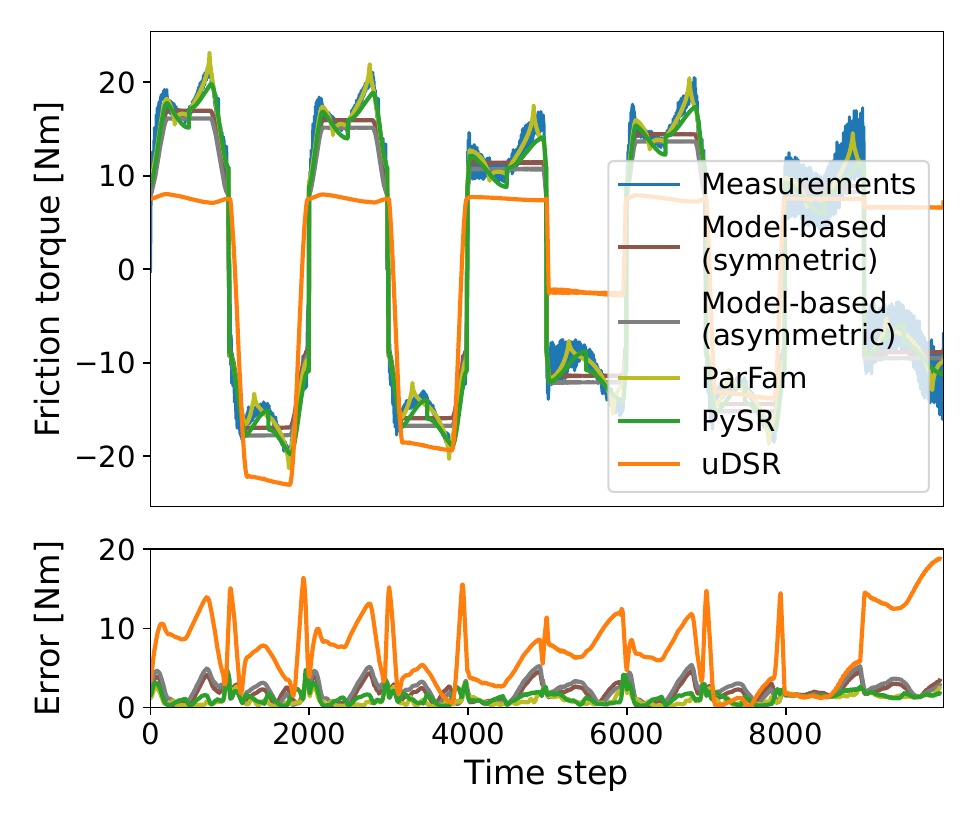}}% \\
    \subfloat[Joint 4 (Error plot is cut at 5Nm in the bottom diagram)]{\includegraphics[width=0.5\linewidth]{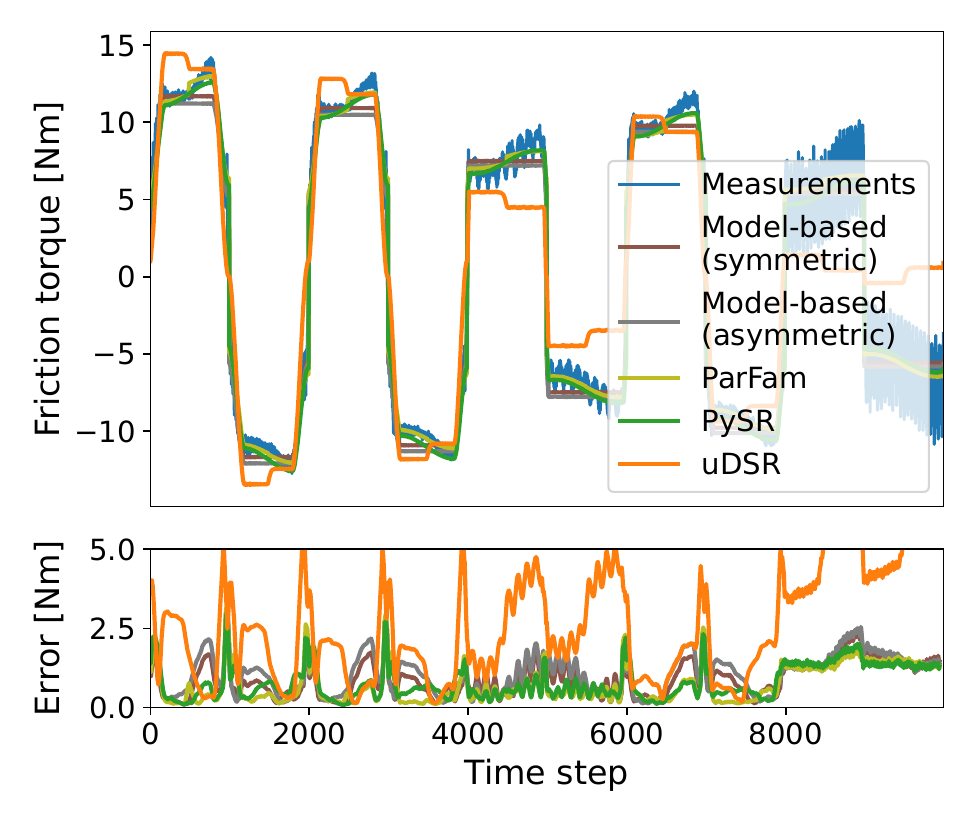}}
    % \end{minipage}
    \caption{Friction estimation on Dataset A incorporating the dependence on velocity and gravitational torque is in the top diagrams of (a) and (b), while the bottom subplots shows the error signals.}
    \label{fig:dataset-a-with-load-dependy-joint-2-only-test-data}
\end{figure}

We randomly select 80 of the 90 movements with constant velocities as training data for the SR algorithms and subsequently subsample the dataset to ensure an equal number of measurements for each movement. The SR algorithms receive $\dot{q}$, $\sgn(\dot{q})$, $\tau_g$, and $\sgn(\tau_g)$ as inputs.
Fig.~\ref{fig:dataset-a-with-load-dependy-joint-2-only-test-data} presents the predictions of classical model-based approaches, which ignore load dependency, and the SR algorithms, which incorporate it, compared to the measurements and the error at each time step for both joints. For better visualization, the error plot for Joint 4 is clipped at 5Nm. The results show that ParFam and PySR significantly outperform the classical models, while uDSR struggles to learn an accurate estimate, as also reflected in Table~\ref{tab:dataset-a-w-load-dependency}.
Given that the algorithms now utilize four-dimensional input, the resulting formulas are more complex. Consequently, due to space constraints, we omit the explicit formulas in Table~\ref{tab:dataset-a-w-load-dependency}. Instead, we present only the most complex but also accurate formula (as shown in Table~\ref{tab:dataset-a-w-load-dependency}), which was derived by ParFam as 

\begin{equation} \label{eq:dataset-a-load-dependency-parfam}
    \scriptstyle
    \begin{split}
    \hat{\tau}_f =  
    & - 1.12 \dot{q}^{2} + 0.01 \dot{q} \tau_g + 1.095 |\dot{q}|
     - 0.01 \dot{q} \sgn(\tau_g) - 14.87 \dot{q} + 0.026 |\tau_g| - 0.089 \tau_g 
    - 0.154 \sgn(\dot{q}) \sgn(\tau_g) \\
    & + 7.162 \sgn(\tau_g) + 0.532
     + \big(- 0.663 \dot{q} + 0.014 \tau_g + 1.64 \sgn(\dot{q}) - 4.245 \sgn(\tau_g)\big. 
     - \big. 10.159 \sgn(\dot{q}) - 0.226\big) \\
    & \cdot \sqrt{\left|{0.167 \dot{q} - 0.076 \tau_g - 0.467 \sgn(\dot{q}) + 2.687 \sgn(\tau_g)}\right|}.
\end{split}
\end{equation}

While this formula is more complex than previous ones, it shows that ParFam and PySR are both able to compute reasonable results that can be interpreted and implemented for real-time execution. Equation~(\ref{eq:dataset-a-load-dependency-parfam}) shows a decent approximation to the physical friction behavior and noticeably captures the four-quadrant effect of the friction torque. This effect is also known as power-quadrant effect, for more details the reader is referred to \citep{scholl2024friction,tadese2021passivity}.     

% \begin{figure}[t]
%     \centering
%     \subfloat[Joint 2]{\includegraphics[width=0.95\linewidth]{Images/dataset_a_with_load_depedency_joint_2.pdf}}\\
%     \subfloat[Joint 4 (Error plot is cut at 5Nm in the bottom diagram)]{\includegraphics[width=0.95\linewidth]{Images/dataset_a_with_load_depedency_joint_4.pdf}}
%     \caption{Friction estimation on Dataset A incorporating the dependence on velocity and gravitational torque is in the top diagrams of (a) and (b), while the bottom subplots shows the error signals.}
%     \label{fig:dataset-a-with-load-dependy-joint-2-only-test-data}
% \end{figure}

\begin{table}[t]
    \centering
    \begin{minipage}[c][0.21\textheight][c]{\columnwidth}
        \centering
        \small % or \footnotesize if space is really tight
        \caption{Results on Dataset A with load dependency}
    \label{tab:dataset-a-w-load-dependency}
        \begin{tabular}{|c|c|c|c|c|}
            \hline
             & \multicolumn{2}{c|}{Estimation error [Nm]} & \multicolumn{2}{c|}{Complexity [-]} \\ 
             & Joint 2 & Joint 4 & Joint 2 & Joint 4\\ \hline
            \makecell{Model-based \\ (symmetric)} & 3.663 & 0.914 & 20 & 20 \\ \hline
            \makecell{Model-based \\ (asymmetric)} & 4.073 & 1.068 & 40 & 40 \\ \hline
            ParFam & 1.025 & 0.724 & 63 & 32 \\ \hline
            PySR & 1.294 & 0.800 & 13 & 8 \\ \hline
            uDSR & 7.168 & 2.671 & 12 & 12 \\ \hline
        \end{tabular}
    \end{minipage}
\end{table}

% \begin{table}
%     \caption{Results on Dataset A with load dependency}
%     \centering
%     \begin{tabular}{|c|c|c|c|c|}
%     \hline
%          & \multicolumn{2}{c|}{Estimation error [Nm]} & \multicolumn{2}{c|}{Complexity [-]} \\ 
%          & Joint 2 & Joint 4 & Joint 2 & Joint 4\\ \hline
%         \makecell{Model-based \\ (symmetric)} & 3.663 & 0.914 & 20 & 20 \\ \hline
%         \makecell{Model-based \\ (asymmetric)}& 4.073 & 1.068 & 40 & 40 \\ \hline
%         ParFam & 1.025 & 0.724 & 63 & 32 \\ \hline
%         PySR & 1.294 & 0.800 & 13 & 8 \\ \hline
%         uDSR & 7.168 & 2.671 & 12 & 12 \\ \hline
%     \end{tabular}
%     \label{tab:dataset-a-w-load-dependency}
% \end{table}

\subsection{Simultaneous motion and external torque estimation} \label{sec:exp-dynamic-dataset}
To conclude our experiments, we explore the feasibility of applying the learned friction model to estimate external torque in new, more complex scenarios. For this, we evaluate the performance of the learned models on Dataset B. Since all joints are moved simultaneously during the measurement of this dataset, the friction model $\hat{\tau}_f$ learned before is slightly off, which is why we use a single movement without external torque to adapt the model. To this end, we learn $\hat{\tau}_{f,add}$ such that $\hat{\tau}_{f}+\hat{\tau}_{f,add}\approx\tau_f$. To prevent the behavior modeled by $\hat{\tau}_{f}$ from being largely overwritten by $\hat{\tau}_{f,add}$, we restrict $\hat{\tau}_{f,add}$ to depend only on $\tau_g$, $\sgn(\tau_g)$, and $\sgn(\dot{q})$, while excluding any direct dependence on $\dot{q}$.

\begin{figure}
    \centering
    \subfloat[\label{fig:dataset-b-asymmetric}Predicted friction torque (Error plot is cut at 10Nm)]{\includegraphics[width=0.5\linewidth]{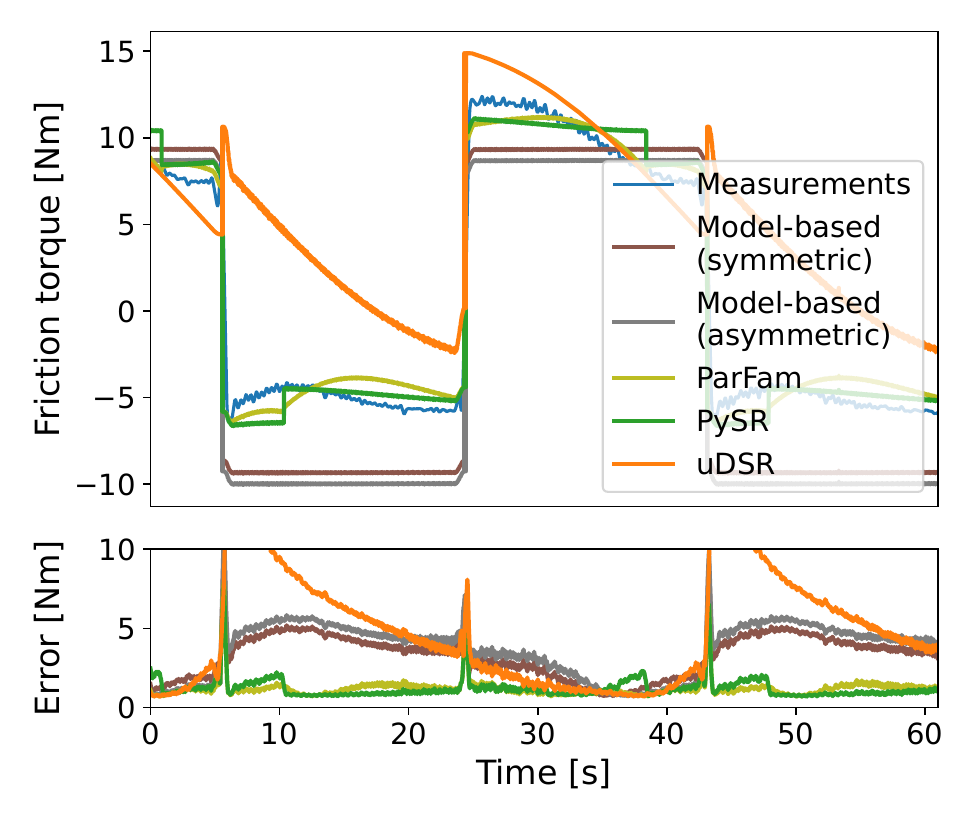}}\\
    \subfloat[\label{fig:dataset-b-asymmetric-external-torque}Predicted external torque]{\includegraphics[width=0.5\linewidth]{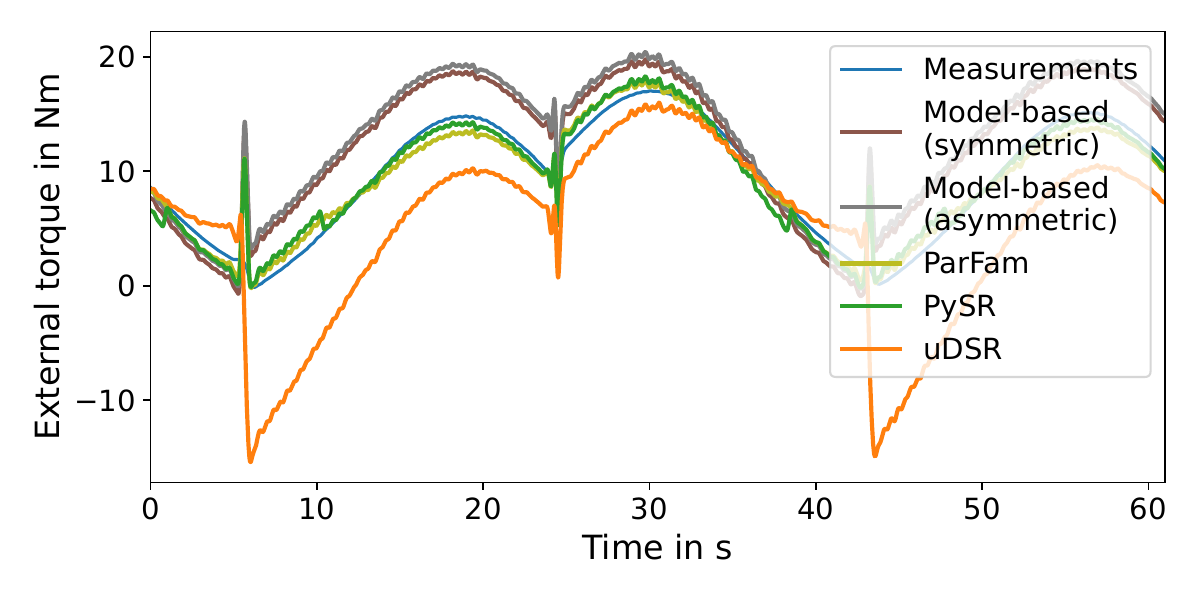}}\\
    \caption{Friction and external torque prediction on the subset of Dataset B external load on Joint~2.}
    \label{fig:dataset-b-load}
\end{figure}

Fig.~\ref{fig:dataset-b-asymmetric} shows the estimates from different models and the corresponding errors at each time step on Joint 2, with errors clipped at 10Nm for visualization purposes. Table~\ref{tab:dataset-b} summarizes the mean errors with and without external loads, as well as the learned formulas and their complexities. The results demonstrate that the estimates by ParFam and PySR outperform the classical model-based approaches, while uDSR fails to learn a good approximation, which was expected since its base model on Dataset A has already been wrong. Interestingly, the formula learned by ParFam shows a direction dependency w.r.t.~velocity and load torque that follows the physical friction behavior. These directional effects are associated mainly with the HD gear and can be modeled conventionally using a four-quadrant approach \citep{tadese2021passivity}.

In Fig.~\ref{fig:dataset-b-asymmetric-external-torque} the friction estimates are then used to predict the external torque, which we compare against measurements from external torque sensors. This demonstrates the potential of friction formulas found by ParFam and PySR for external torque estimation, reducing the error from over 3Nm (using model-based approaches) to approximately 1Nm on Joint 2, and from 1.5--1.6Nm to 0.9--1.4Nm for Joint 4.

\begin{table*}
\centering
\scriptsize
    \centering
    \caption{Results on Dataset B: friction estimation with/without load, complexity, and the computed formular $\hat{\tau}_f$ }
    \label{tab:dataset-b}
    \begin{threeparttable}
    \begin{tabular}{|c|c|c|c|c|c|c|c|}
        \hline
         & \multicolumn{2}{c|}{\makecell{Without \\ load [Nm]}} & \multicolumn{2}{c|}{\makecell{With \\ load [Nm]}} & \multicolumn{2}{c|}{\makecell{Complexity [-]\\ (adaption only)}} & Formula (adaption only) \\
         & Joint 2 & Joint 4 & Joint 2 & Joint 4 & Joint 2 & Joint 4 & Joint 2 \\ \hline
        \makecell{Model-based \\ (symmetric)} & 3.663 & 1.473 & 3.171 & 1.628 & 20 & 20 & (no adaption) \\ \hline
        \makecell{Model-based \\ (asymmetric)} & 4.073 & 1.318 & 3.550 & 1.648 & 40 & 40 & (no adaption) \\ \hline
        ParFam & 0.611 & 0.742 & 1.007 & 1.189 & 20 & 11 & \makecell[l]{
            $\begin{aligned}
                \hat{\tau}_f = &- 0.012 g^{2} + 0.016 g \sgn(\dot{q}) + 0.176 |g|  - \\  & 0.169 g + 0.712 \sgn(\dot{q}) + 1.036
            \end{aligned}$
        }\\ \hline
        PySR & 0.911 & 0.936 & 0.953 & 1.428 &  3 & 2 & $\hat{\tau}_f=\left(\sgn(g) - 0.9893261\right)^{2}$\\ \hline
        uDSR & 4.824 & 2.238 & 4.502 & 1.636 &  5 & 5 & $\hat{\tau}_f=0.330 \sgn(\dot{q}) \left(g - 1\right)$ \\ \hline    \end{tabular}

        \begin{tablenotes}
        \footnotesize
        \item[*] The estimation errors with/without load denote the mean absolute errors of the friction estimation in Nm.
        \item[*] For brevity, we present the complexity of the computed formulas for Joints~2 and 4, and the computed formula for $\hat{\tau}_f$ for Joint~2.
        \end{tablenotes}
    \end{threeparttable}
    \vspace{-0.4cm}
\end{table*}

\section{Conclusion} \label{sec:conclusion}
In this study, we addressed the longstanding challenge of accurately modeling friction torque in robotic joints by leveraging the symbolic regression (SR) approach as an alternative to traditional model-based and data-driven methods. Our findings demonstrate that SR can generate interpretable, mathematically concise formulas that not only rival but surpass the accuracy of model-based methods. Moreover, SR's flexibility allows for seamless incorporation of additional dependencies, such as load effects, making it suitable for diverse robotic applications.
The proposed approach bridges the gap between robustness and interpretability, offering a reliable and efficient solution for estimating friction torque in robotic systems. This makes SR particularly valuable for control and physical interaction scenarios, where trustworthiness and transparency are essential.

\section*{Acknowledgements}
P.S. and G. K. acknowledge support by the project "Genius Robot" (01IS24083), funded by the Federal Ministry of Education and Research (BMBF) and additional support from LMUexcellent, funded by the Federal Ministry of Education and Research (BMBF) and the Free State of Bavaria under the Excellence Strategy of the Federal Government and the L\"ander as well as by the Hightech Agenda Bavaria. 
%Bibliography
% \bibliographystyle{unsrt}  
\bibliographystyle{abbrvnat}
\bibliography{references}  

\providecommand{\noopsort}[1]{}\providecommand{\singleletter}[1]{#1}%
\begin{thebibliography}{39}
\providecommand{\natexlab}[1]{#1}
\providecommand{\url}[1]{\texttt{#1}}
\expandafter\ifx\csname urlstyle\endcsname\relax
  \providecommand{\doi}[1]{doi: #1}\else
  \providecommand{\doi}{doi: \begingroup \urlstyle{rm}\Url}\fi

\bibitem[Armstrong-Helouvry(2012)]{armstrong2012}
B.~Armstrong-Helouvry.
\newblock \emph{Control of machines with friction}, volume 128.
\newblock Springer Science \& Business Media, 2012.

\bibitem[Augusto and Barbosa(2000)]{augusto2000symbolic}
D.~A. Augusto and H.~J. Barbosa.
\newblock Symbolic regression via genetic programming.
\newblock In \emph{Proceedings. Vol.1. Sixth Brazilian Symposium on Neural Networks}, pages 173--178. IEEE, 2000.
\newblock \doi{10.1109/SBRN.2000.889734}.

\bibitem[Bertsimas and Tsitsiklis(1993)]{bertsimas1993simulated}
D.~Bertsimas and J.~Tsitsiklis.
\newblock Simulated annealing.
\newblock \emph{Statistical science}, 8\penalty0 (1):\penalty0 10--15, 1993.

\bibitem[Bittencourt and Axelsson(2013)]{bittencourt2013modeling}
A.~C. Bittencourt and P.~Axelsson.
\newblock Modeling and experiment design for identification of wear in a robot joint under load and temperature uncertainties based on friction data.
\newblock \emph{IEEE/ASME transactions on mechatronics}, 19\penalty0 (5):\penalty0 1694--1706, 2013.

\bibitem[Bittencourt and Gunnarsson(2012)]{bittencourt2012static}
A.~C. Bittencourt and S.~Gunnarsson.
\newblock Static friction in a robot joint—modeling and identification of load and temperature effects.
\newblock 2012.

\bibitem[Brindle(1980)]{brindle1980genetic}
A.~Brindle.
\newblock Genetic algorithms for function optimization.
\newblock 1980.

\bibitem[Brunton et~al.(2016)Brunton, Proctor, and Kutz]{brunton2016}
S.~L. Brunton, J.~L. Proctor, and J.~N. Kutz.
\newblock Discovering governing equations from data by sparse identification of nonlinear dynamical systems.
\newblock \emph{Proceedings of the National Academy of Sciences}, 113\penalty0 (15):\penalty0 3932--3937, 2016.
\newblock \doi{10.1073/pnas.1517384113}.

\bibitem[Canudas~de Wit et~al.(1995)Canudas~de Wit, Olsson, Astrom, and Lischinsky]{Canudas1995}
C.~Canudas~de Wit, H.~Olsson, K.~Astrom, and P.~Lischinsky.
\newblock A new model for control of systems with friction.
\newblock \emph{IEEE Transactions on Automatic Control}, 40\penalty0 (3):\penalty0 419--425, 1995.
\newblock \doi{10.1109/9.376053}.

\bibitem[Cranmer(2023)]{cranmer2023interpretable}
M.~Cranmer.
\newblock Interpretable machine learning for science with pysr and symbolicregression. jl.
\newblock \emph{arXiv preprint arXiv:2305.01582}, 2023.

\bibitem[De~Santis et~al.(2008)De~Santis, Siciliano, De~Luca, and Bicchi]{de2008atlas}
A.~De~Santis, B.~Siciliano, A.~De~Luca, and A.~Bicchi.
\newblock An atlas of physical human--robot interaction.
\newblock \emph{Mechanism and Machine Theory}, 43\penalty0 (3):\penalty0 253--270, 2008.

\bibitem[Guo et~al.(2019)Guo, Pan, and Yu]{Guo2019}
K.~Guo, Y.~Pan, and H.~Yu.
\newblock Composite learning robot control with friction compensation: A neural network-based approach.
\newblock \emph{IEEE Transactions on Industrial Electronics}, 66\penalty0 (10):\penalty0 7841--7851, 2019.
\newblock \doi{10.1109/TIE.2018.2886763}.

\bibitem[He et~al.(2016)He, Zhang, Ren, and Sun]{he2016deep}
K.~He, X.~Zhang, S.~Ren, and J.~Sun.
\newblock Deep residual learning for image recognition.
\newblock In \emph{Proceedings of the IEEE conference on computer vision and pattern recognition}, pages 770--778, 2016.

\bibitem[Huang and Tan(2012)]{Huang2012}
S.~Huang and K.~K. Tan.
\newblock Intelligent friction modeling and compensation using neural network approximations.
\newblock \emph{IEEE Transactions on Industrial Electronics}, 59\penalty0 (8):\penalty0 3342--3349, 2012.
\newblock \doi{10.1109/TIE.2011.2160509}.

\bibitem[Iskandar and Wolf(2019)]{iskandar2019}
M.~Iskandar and S.~Wolf.
\newblock Dynamic friction model with thermal and load dependency: modeling, compensation, and external force estimation.
\newblock In \emph{2019 International Conference on Robotics and Automation (ICRA)}, pages 7367--7373. IEEE, 2019.

\bibitem[Iskandar et~al.(2021)Iskandar, Eiberger, Albu-Sch{\"a}ffer, De~Luca, and Dietrich]{iskandar2021}
M.~Iskandar, O.~Eiberger, A.~Albu-Sch{\"a}ffer, A.~De~Luca, and A.~Dietrich.
\newblock Collision detection, identification, and localization on the dlr sara robot with sensing redundancy.
\newblock In \emph{2021 IEEE International Conference on Robotics and Automation (ICRA)}, pages 3111--3117. IEEE, 2021.

\bibitem[Iskandar et~al.(2023)Iskandar, Ott, Albu-Sch{\"a}ffer, Siciliano, and Dietrich]{iskandar2023hybrid}
M.~Iskandar, C.~Ott, A.~Albu-Sch{\"a}ffer, B.~Siciliano, and A.~Dietrich.
\newblock Hybrid force-impedance control for fast end-effector motions.
\newblock \emph{IEEE Robotics and Automation Letters}, 2023.

\bibitem[Iskandar et~al.(2024)Iskandar, Albu-Sch{\"a}ffer, and Dietrich]{iskandar2024intrinsic}
M.~Iskandar, A.~Albu-Sch{\"a}ffer, and A.~Dietrich.
\newblock Intrinsic sense of touch for intuitive physical human-robot interaction.
\newblock \emph{Science Robotics}, 9\penalty0 (93):\penalty0 eadn4008, 2024.

\bibitem[Johanastrom and Canudas-De-Wit(2008)]{johanastrom2008revisiting}
K.~Johanastrom and C.~Canudas-De-Wit.
\newblock Revisiting the lugre friction model.
\newblock \emph{IEEE Control systems magazine}, 28\penalty0 (6):\penalty0 101--114, 2008.

\bibitem[Kamienny et~al.(2022)Kamienny, d'Ascoli, Lample, and Charton]{kamienny2022end}
P.~Kamienny, S.~d'Ascoli, G.~Lample, and F.~Charton.
\newblock End-to-end symbolic regression with transformers.
\newblock In \emph{Advances in Neural Information Processing Systems}, 2022.

\bibitem[{Kemal Cılız} and Tomizuka(2007)]{Ciliz2007}
M.~{Kemal Cılız} and M.~Tomizuka.
\newblock Friction modelling and compensation for motion control using hybrid neural network models.
\newblock \emph{Engineering Applications of Artificial Intelligence}, 20\penalty0 (7):\penalty0 898--911, 2007.
\newblock ISSN 0952-1976.
\newblock \doi{https://doi.org/10.1016/j.engappai.2006.12.007}.

\bibitem[La~Cava et~al.(2021)La~Cava, Orzechowski, Burlacu, de~Fran{\c{c}}a, Virgolin, Jin, Kommenda, and Moore]{la2021contemporary}
W.~G. La~Cava, P.~Orzechowski, B.~Burlacu, F.~O. de~Fran{\c{c}}a, M.~Virgolin, Y.~Jin, M.~Kommenda, and J.~H. Moore.
\newblock Contemporary symbolic regression methods and their relative performance.
\newblock In J.~Vanschoren and S.~Yeung, editors, \emph{Proceedings of the Neural Information Processing Systems Track on Datasets and Benchmarks}, 2021.

\bibitem[Lakatos et~al.(2013)Lakatos, G{\"o}rner, Petit, Dietrich, and Albu-Sch{\"a}ffer]{lakatos2013modally}
D.~Lakatos, M.~G{\"o}rner, F.~Petit, A.~Dietrich, and A.~Albu-Sch{\"a}ffer.
\newblock A modally adaptive control for multi-contact cyclic motions in compliantly actuated robotic systems.
\newblock In \emph{2013 IEEE/RSJ International Conference on Intelligent Robots and Systems}, pages 5388--5395. IEEE, 2013.

\bibitem[Landajuela et~al.(2022)Landajuela, Lee, Yang, Glatt, Santiago, Aravena, Mundhenk, Mulcahy, and Petersen]{Landajuela2022usdr}
M.~Landajuela, C.~S. Lee, J.~Yang, R.~Glatt, C.~P. Santiago, I.~Aravena, T.~Mundhenk, G.~Mulcahy, and B.~K. Petersen.
\newblock A unified framework for deep symbolic regression.
\newblock In S.~Koyejo, S.~Mohamed, A.~Agarwal, D.~Belgrave, K.~Cho, and A.~Oh, editors, \emph{Advances in Neural Information Processing Systems}, volume~35, pages 33985--33998. Curran Associates, Inc., 2022.

\bibitem[Linderoth et~al.(2013)Linderoth, Stolt, Robertsson, and Johansson]{Linderoth2013}
M.~Linderoth, A.~Stolt, A.~Robertsson, and R.~Johansson.
\newblock Robotic force estimation using motor torques and modeling of low velocity friction disturbances.
\newblock In \emph{2013 IEEE/RSJ International Conference on Intelligent Robots and Systems}, pages 3550--3556, 2013.
\newblock \doi{10.1109/IROS.2013.6696862}.

\bibitem[Liu et~al.(2021)Liu, Wang, and Wang]{liu2021sensorless}
S.~Liu, L.~Wang, and X.~V. Wang.
\newblock Sensorless force estimation for industrial robots using disturbance observer and neural learning of friction approximation.
\newblock \emph{Robotics and Computer-Integrated Manufacturing}, 71:\penalty0 102168, 2021.

\bibitem[Martius and Lampert(2017)]{martius2016extrapolation}
G.~Martius and C.~H. Lampert.
\newblock Extrapolation and learning equations.
\newblock In \emph{5th International Conference on Learning Representations, Workshop Track Proceedings}, 2017.

\bibitem[Mundhenk et~al.(2021)Mundhenk, Landajuela, Glatt, Santiago, Faissol, and Petersen]{mundhenk2021symbolic}
T.~N. Mundhenk, M.~Landajuela, R.~Glatt, C.~P. Santiago, D.~M. Faissol, and B.~K. Petersen.
\newblock Symbolic regression via deep reinforcement learning enhanced genetic programming seeding.
\newblock In \emph{Advances in Neural Information Processing Systems}, volume~34, pages 24912--24923, 2021.

\bibitem[Petersen et~al.(2021)Petersen, Landajuela, Mundhenk, Santiago, Kim, and Kim]{petersen2019deep}
B.~K. Petersen, M.~Landajuela, T.~N. Mundhenk, C.~P. Santiago, S.~Kim, and J.~T. Kim.
\newblock Deep symbolic regression: Recovering mathematical expressions from data via risk-seeking policy gradients.
\newblock In \emph{9th International Conference on Learning Representations, {ICLR} 2021}, 2021.

\bibitem[Real et~al.(2019)Real, Aggarwal, Huang, and Le]{real2019regularized}
E.~Real, A.~Aggarwal, Y.~Huang, and Q.~V. Le.
\newblock Regularized evolution for image classifier architecture search.
\newblock In \emph{Proceedings of the aaai conference on artificial intelligence}, volume~33, pages 4780--4789, 2019.

\bibitem[Sandi et~al.(2023)Sandi, Vedran, Jasna, and Car]{sandi2023determining}
B.~{\v{S}}. Sandi, M.~Vedran, P.-O. Jasna, and Z.~Car.
\newblock Determining normalized friction torque of an industrial robotic manipulator using the symbolic regression method.
\newblock \emph{Industry 4.0}, 8\penalty0 (1):\penalty0 21--24, 2023.

\bibitem[Scholl et~al.(2024)Scholl, Iskandar, Wolf, Lee, Bacho, Dietrich, Albu-Schäffer, and Kutyniok]{scholl2024friction}
P.~Scholl, M.~Iskandar, S.~Wolf, J.~Lee, A.~Bacho, A.~Dietrich, A.~Albu-Schäffer, and G.~Kutyniok.
\newblock Learning-based adaption of robotic friction models.
\newblock \emph{Robotics and Computer-Integrated Manufacturing}, 89:\penalty0 102780, 2024.
\newblock ISSN 0736-5845.
\newblock \doi{https://doi.org/10.1016/j.rcim.2024.102780}.

\bibitem[Scholl et~al.(2025)Scholl, Bieker, Hauger, and Kutyniok]{scholl2023parfam}
P.~Scholl, K.~Bieker, H.~Hauger, and G.~Kutyniok.
\newblock Parfam--(neural guided) symbolic regression via continuous global optimization.
\newblock In \emph{The Thirteenth International Conference on Learning Representations}, 2025.

\bibitem[Selmic and Lewis(2002)]{Selmic2002}
R.~Selmic and F.~Lewis.
\newblock Neural-network approximation of piecewise continuous functions: application to friction compensation.
\newblock \emph{IEEE Transactions on Neural Networks}, 13\penalty0 (3):\penalty0 745--751, 2002.
\newblock \doi{10.1109/TNN.2002.1000141}.

\bibitem[Tadese et~al.(2021)Tadese, Yumbla, Yi, Lee, Park, and Moon]{tadese2021passivity}
M.~A. Tadese, F.~Yumbla, J.-S. Yi, W.~Lee, J.~Park, and H.~Moon.
\newblock Passivity guaranteed dynamic friction model with temperature and load correction: Modeling and compensation for collaborative industrial robot.
\newblock \emph{IEEE Access}, 9:\penalty0 71210--71221, 2021.

\bibitem[Udrescu and Tegmark(2020)]{udrescu2020tegmark}
S.-M. Udrescu and M.~Tegmark.
\newblock Ai feynman: A physics-inspired method for symbolic regression.
\newblock \emph{Science Advances}, 6\penalty0 (16):\penalty0 eaay2631, 2020.
\newblock \doi{10.1126/sciadv.aay2631}.

\bibitem[Virtanen~et al.(2020)]{2020SciPy-NMeth}
P.~Virtanen~et al.
\newblock {{SciPy} 1.0: Fundamental Algorithms for Scientific Computing in Python}.
\newblock \emph{Nature Methods}, 17:\penalty0 261--272, 2020.
\newblock \doi{10.1038/s41592-019-0686-2}.

\bibitem[Vogel et~al.(2020)Vogel, Leidner, Hagengruber, Panzirsch, Bauml, Denninger, Hillenbrand, Suchenwirth, Schmaus, Sewtz, et~al.]{vogel2020ecosystem}
J.~Vogel, D.~Leidner, A.~Hagengruber, M.~Panzirsch, B.~Bauml, M.~Denninger, U.~Hillenbrand, L.~Suchenwirth, P.~Schmaus, M.~Sewtz, et~al.
\newblock An ecosystem for heterogeneous robotic assistants in caregiving: Core functionalities and use cases.
\newblock \emph{IEEE Robotics \& Automation Magazine}, 28\penalty0 (3):\penalty0 12--28, 2020.

\bibitem[Wales and Doye(1997)]{Wales1997GlobalOB}
D.~J. Wales and J.~P.~K. Doye.
\newblock Global optimization by basin-hopping and the lowest energy structures of lennard-jones clusters containing up to 110 atoms.
\newblock \emph{Journal of Physical Chemistry A}, 101:\penalty0 5111--5116, 1997.
\newblock \doi{https://doi.org/10.1021/jp970984n}.

\bibitem[Wolf and Iskandar(2018)]{wolf2018}
S.~Wolf and M.~Iskandar.
\newblock Extending a dynamic friction model with nonlinear viscous and thermal dependency for a motor and harmonic drive gear.
\newblock In \emph{2018 IEEE International Conference on Robotics and Automation (ICRA)}, pages 783--790. IEEE, 2018.

\end{thebibliography}
\end{document}